\documentclass[10pt, a4paper]{article}
\usepackage{array}

\usepackage[final]{lrec2026}

\usepackage{times}
\usepackage{latexsym}
\usepackage{multirow}
\usepackage{makecell}
\usepackage{graphics}
\usepackage{booktabs}
\usepackage{hyperref}
\usepackage{lineno}

\usepackage{array,booktabs,multirow}
\usepackage[T1]{fontenc}
\usepackage[utf8]{inputenc}

\usepackage{microtype}

\usepackage{inconsolata}

\usepackage{graphicx}

\usepackage{pifont}
\usepackage{subcaption}

\newcommand{\cmark}{\ding{51}}
\newcommand{\xmark}{\ding{55}}

\title{\includegraphics[height=2.5ex]{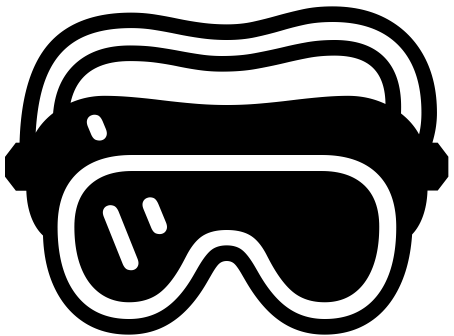}~The Chronicles of \textsc{RiDiC}: Generating Datasets with Controlled Popularity Distribution for Long-form Factuality Evaluation}

\name{\begin{tabular}{c}
Pavel Braslavski\textsuperscript{1,2}, 
Dmitrii Iarosh\textsuperscript{3}, 
Nikita Sushko\textsuperscript{4}, 
Andrey Sakhovskiy\textsuperscript{4,6}, \\
Vasily Konovalov\textsuperscript{5}, 
Elena Tutubalina\textsuperscript{1,5,6}, 
Alexander Panchenko\textsuperscript{4,5}
\end{tabular}}

\address{
\textsuperscript{1}HSE University, \textsuperscript{2}Ural Federal University, \textsuperscript{3}ITMO University, \textsuperscript{4}Skoltech,
\textsuperscript{5}AIRI,
\textsuperscript{6}Sber AI\\
\texttt{pbras@yandex.ru},\texttt{\{andrei.sakhovskii, a.panchenko\}@skol.tech}  
}

\abstract{
We present a configurable pipeline 
for generating multilingual sets of entities with specified characteristics, such as domain, geographical location and \textit{popularity}, using data from Wikipedia and Wikidata. 
These datasets are intended for evaluating the factuality of LLMs' long-form generation,
thereby complementing evaluation based on short-form QA datasets.
We present the \textsc{RiDiC} dataset as an example of this approach. \textsc{RiDiC} contains 3,000 entities from three domains -- rivers, natural disasters, and car models -- spanning different popularity tiers. Each entity is accompanied by its geographical location, English and Chinese names (if available) and relevant English and Chinese Wikipedia content, which is used to evaluate LLMs' responses.  
Generations about \textsc{RiDiC} entities were obtained from three LLMs
in English and Chinese. These were then evaluated using a third-party factuality checker, which showed that entities from our dataset caused even frontier models to hallucinate. 
To facilitate the evaluation of LLMs' long-form factuality in multiple languages, the code, data, and generation/evaluation scripts have been released.  
 \\ \newline  }

\begin{document}
\maketitleabstract
% \maketitle
% \begin{abstract}
% We present a pipeline and associated code for generating sets of entities with specified characteristics, such as domain, geography, and \textit{popularity}, using data from both Wikipedia and Wikidata. The pipeline is highly configurable and multilingual by design, and implements a variety of entity popularity measures.  
% These datasets are aimed at evaluating the factuality of LLMs' long-form generations in different domains and popularity tiers, thus complementing evaluation based on QA datasets with ground-truth information. 
% We present the RiDiC dataset as an example of this approach. RiDiC contains 1,000 entities from three classes -- rivers, natural disasters, and car models -- from different popularity tiers. Each entity is accompanied by its location, English and Chinese names (if available) and relevant English and Chinese Wikipedia content, which is used to evaluate LLMs' responses.  
% Generations about RiDiC entities were obtained from two open medium-sized LLMs (LLama 8B and Qwen 7B) and a closed frontier model (GPT-5) in English and Chinese. We then evaluated these using a third-party factuality checker, FactOWL, with Wikipedia as the source of evidence. \pb{Experimental results show that ...} 
%1. What did we do
%2. Why did we do it
%3. How did we do it
%4. What did we find
%5. What do we think this means

% \end{abstract}

\section{Introduction}

For many users, LLMs have become the go-to tool for information-seeking tasks.
LLMs provide coherent answers, eliminating the need to sift through numerous documents to find, analyze, and organize information. However, LLM responses can contain factual errors, i.e., information that contradicts knowledge accumulated in reliable sources, such as Wikipedia, dictionaries, and textbooks. These errors can be especially critical in domains such as health, law, finance, and security. Factual errors are more difficult for users to detect than input- or context-conflicting hallucinations because responses look plausible, contain no obvious contradictions, and are expressed confidently~\cite{augenstein2024factuality}.

Various approaches exist to enhance the factuality of LLMs, such as using external knowledge (RAG) and improving internal factuality with pre- and post-training techniques~\cite{wang2025survey}. However, in order to evaluate LLMs and track their progress, factuality evaluation benchmarks are necessary. Due to the multitasking nature and wide range of LLM applications, creating a universal benchmark is unrealistic; instead, dedicated datasets are built to evaluate different aspects of LLMs' factuality. For example, MMLU~\cite{mmlu} is designed to probe both the world knowledge and problem-solving abilities of LLMs across a wide range of tasks and domains. TruthfulQA~\cite{lin-etal-2022-truthfulqa} allows one to check whether LLMs have learned common misconceptions or false beliefs during training. FreshQA~\cite{vu-etal-2024-freshllms} is a dynamic benchmark comprising both evergreen and fast-changing questions designed to evaluate LLMs' up-to-date world knowledge. 
% Dynamic evaluation suites FreshQA and LiveBench~\cite{livebench} address the problem of contamination of open static benchmarks. 

LLM benchmarks also differ in format. 
For instance, MMLU is a multiple-choice dataset, whereas SimpleQA~\cite{simpleqa} contains questions and short answers. 
One advantage of these formats is that it is relatively easy to match an LLM answer with the provided correct answer. However, it is crucial to evaluate \textit{long-form} generation because most real-world user requests demand comprehensive and coherent responses rather than isolated facts or short answers.
% the biggest LLMs' advantage, which attracts users 
% , including in information retrieval tasks, 
% is their ability to provide a coherent high-quality text as a response. 

At the same time, it is unclear how short-answer factuality correlates with the ability to produce longer narratives containing numerous facts~\cite{islam2025curiouscasefactualmisalignment}.
% As the authors write, SimpleQA ``whether the ability to provide factual short answers correlates with the ability to write lengthy responses filled with numerous facts remains an open research question.'' 
FActScore~\cite{min-etal-2023-factscore}, a collection of person names and an eponymous evaluation framework, was seminal work on long-form factuality evaluation. Notably, the composition of the dataset is guided by the persons' popularity. As previous studies showed, LLMs' factual knowledge is strongly correlated with the popularity of pertinent entities~\cite{kandpal2023large,mallen-etal-2023-trust}. %Other long-form evaluation datasets -- LongFact and WildHallucinations -- are composed without accounting for items’ popularity.
%There is a special class of benchmarks comprising of items with varying frequency/popularity, such as question answering datasets PopQA (see Section~\ref{sec:relwork}). As \citet{kandpal2023large} showed, LLMs' factual knowledge is strongly correlated with the frequency of pertinent entities in the pre-training corpus.   

\begin{table*}[th!]
    \centering
    \small
    \begin{tabular}{llclr}
\hline
\textbf{Dataset} & \textbf{Format} & \textbf{Code} & \textbf{Popularity} & \textbf{Size}\\
\hline
EntityQuestions~\cite{sciavolino-etal-2021-simple} & QA & \xmark & Wikipedia links\textsuperscript{\textdagger} & 24k\\
PopQA~\cite{min-etal-2023-factscore} & QA & \xmark & Wikipedia pageviews & 14k \\
FActScore~\cite{inetal2023factscore} & Freeform & \xmark & Wikipedia pageviews/frequency & 183\\
TriviaQA/NQ~\cite{kandpal2023large}& QA & \xmark & Corpus frequency\textsuperscript{\textdagger}  & 100k+\\
Head-to-Tail~\cite{sun-etal-2024-head} & QA & \cmark & Wikipedia pageviews/density  & 18k\\
WildHallucinations~\cite{zhao2024wildhallucinations} & Freeform & \xmark & Perplexity\textsuperscript{\textdagger} & 8k\\
WiTQA~\cite{maekawa-etal-2024-retrieval} & QA & \xmark & Wikidata triples & 14k\\
LongFact~\cite{wei-2024-longform} & Freeform & \xmark & -- & 2.2k\\
LTGen~\cite{ltgen} & QA  & \cmark &  Wikipedia pageviews  & 19k\\
% FACTOR & Factual Corpora (Wiki, News) & Auto-transform corpus: generate false variations for true statements via InstructGPT & None & Multiple Choice (Likelihood) & General (Wiki, News), Expert & 4.2k\\
% MKJ & UMLS & Extract triples, apply templates, entity substitution for T/F medical statements & PubMed frequency (post hoc) & True/False & Medicine & 3k\\
% \textbf{Ours} & freeform & various & various classes & xxx\\
\textbf{RiDiC (this paper)} & Freeform & \cmark & Wikipedia pageviews  & 3k\\
\hline
\end{tabular}
\caption{Datasets with popularity facet (\textsuperscript{\textdagger}\textit{post hoc} popularity analysis). Third column indicates if the code for dataset generation is available.} \label{tab:datasets}
\end{table*}

In this study, we present a flexible pipeline for generating datasets with the desired popularity distribution of their elements. First, we collect entities from a given class on Wikidata and select those with a Wikipedia page. Then, we calculate various popularity metrics based on Wikipedia and Wikidata. Next, we sample entities according to the chosen popularity measure and desired distribution. Finally, we collect Wikipedia data to serve as evidence for fact verification. 
%: the entity's Wikipedia page, Wikipedia search results for the entity, and pages linking to the entity's page. %Optionally, this data can be supplemented with documents from the ``external links'' section of the Wikipedia page and web search results. 

As an example of our approach, we generated \textsc{RiDiC} -- a dataset containing three types of entities: \textbf{Ri}vers, natural \textbf{Di}sasters, and \textbf{C}ar models.
% rivers (natural objects), natural disasters (events), and cars (technological objects). 
The dataset contains 1,000 entities of each type in three popularity tiers (head-torso-tail) based on Wikipedia pageview statistics. We gathered responses from three LLMs in two languages -- English and Chinese -- for these entities and evaluated their factuality. \textsc{RiDiC} can be seen as an extension of FActScore along three dimensions: size, domains, and languages. 
%Moreover, we introduce a highly flexible dataset generation pipeline that allows an easy customization for specific needs.
% \pb{Short overview of results.}

Our contribution is three-fold: 
\begin{enumerate}
    \item We introduced a flexible, multilingual pipeline for generating datasets with controlled entity popularity. This enables a systematic evaluation of long-form factuality in LLMs across domains, geographies, and popularity tiers.
    \item We released \textsc{RiDiC}, a dataset comprising 3,000 entities across three domains (rivers, natural disasters, and car models), to assess long-form factual accuracy in both English and Chinese.
    \item We explored the correlation between entity popularity, location, domain, and factual precision in multilingual LLM long-form outputs. This research provides new insights into long-tail factuality challenges and multilingual evaluation reliability.
\end{enumerate}

We made the code and data freely available.\footnote{\url{https://github.com/s-nlp/ridic}}

\section{Related Work} \label{sec:relwork}

Table~\ref{tab:datasets} summarizes the main characteristics of the datasets with the popularity facet. These datasets differ in format, size, popularity proxy used, and domain. Another important difference is that some datasets have been designed to comply with a given popularity distribution of their elements, while the popularity-related analysis of others was conducted \textit{post hoc}. For example, \citet{kandpal2023large} performed massive entity linking in a pre-training corpus, enabling them to estimate the frequency of individual entities and match them with the popular NaturalQuestions~\cite{kwiatkowski2019natural} and TriviaQA~\cite{joshi-etal-2017-triviaqa} datasets. Most datasets use pageview statistics from corresponding Wikipedia page as a proxy for entity popularity. Alternative approaches include link-based popularity (Head-to-Tail, EntityQuestions), the frequency of the triples containing the entity in Wikidata or another knowledge base (Head-to-Tail, WiTQA), and perplexity of the entity name based on a specific LLM (WildHallucinations).

\begin{figure*}[th!]
    \centering
    \includegraphics[width=0.9\linewidth]{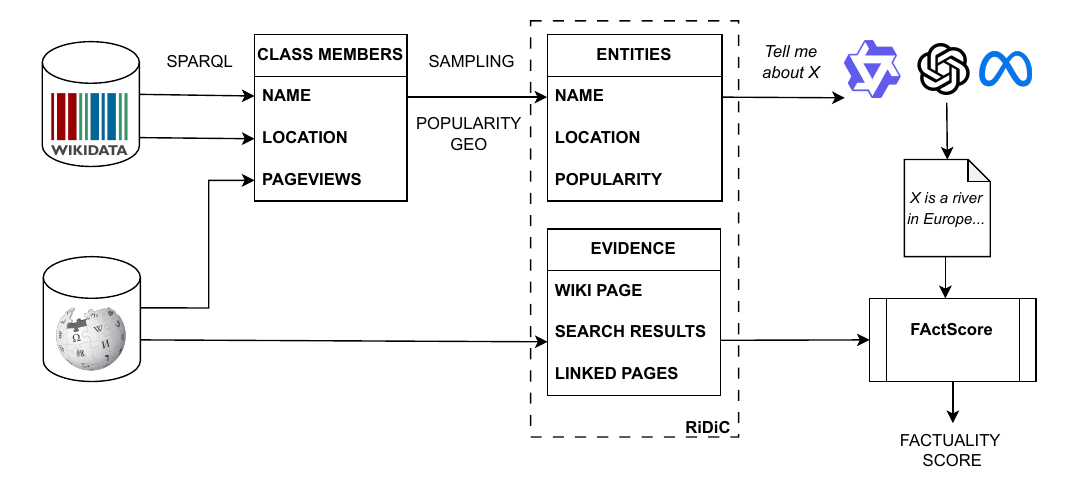}
    \caption{The process of generating a dataset begins with a Wikidata SPARQL query that defines the class of interest. Then, for each class member, attributes and popularity statistics are gathered from Wikidata and Wikipedia. The dataset is formed by sampling the required number of entities with the desired popularity distribution; Wikipedia content is collected as evidence. Once the dataset is complete, the LLMs' generated content about the collected entities is evaluated using a factuality checker.}
    \label{fig:pipeline}
\end{figure*}

The majority of the datasets are based on factoid questions. This format has an obvious advantage: it is relatively simple to compare the obtained answer against the reference answer. The questions come from a web search log (Natural Questions) or are generated based on Wikidata triples using templates (Entity Questions, PopQA, and Head-to-Tail) or LLMs (WiTQA and LTGen). However, the QA format differs from common practical scenarios: users typically ask LLMs more general questions and value detailed responses that reflect various aspects of a problem or entity. The FActScore dataset~\cite{min-etal-2023-factscore} was pioneering work that proposed an approach for evaluating the factuality of long-form LLM generation. The dataset consists of 183 person names sampled based on their popularity, as measured by Wikipedia page views and frequency in a large corpus, as well as geography. 
The disadvantages of this dataset are its modest size and its scope limited to biographies.

Many existing public datasets may have been contaminated by exposure to LLMs during pre-training or fine-tuning. Therefore, a recent trend is to develop an automatic pipeline for generating datasets with desired characteristics instead of creating a static dataset once~\cite{sun-etal-2024-head,maekawa-etal-2024-retrieval}.

Datasets for evaluating long-form generation don't contain specific questions or correct answers. They only contain a set of entities and possibly a prompt template, such as \textit{tell me about X}. Thus, the focus shifts from generating questions to evaluating the LLM's long response. 
The FActScore authors conducted a manual evaluation of responses of three LLMs and proposed an automated pipeline that approximates human scores. This approach has been widely used since then. Subsequent studies have built upon this approach in various ways, including improved extraction of atomic facts, evidence retrieval using web search, and more sophisticated fact verification using multi-step reasoning~\cite{song-etal-2024-veriscore,wei-2024-longform,metropolitansky2025towards}.

There are several studies applying FActScore methodology to LLMs' generations about persons in non-English languages~\cite{kim-etal-2024-analysis,shafayat2024multi,chataigner2024multilingual}. 
These studies demonstrate that generations in higher-resourced languages are more factual. \citet{shafayat2024multi} also showed that generations about people from the Western world are of higher quality, regardless of the language. These studies also indicate that the evaluation is unreliable in less-resourced languages due to lower-quality atomic fact extraction and scarcer knowledge sources that can be used as evidence.
% of multilingual FActScore~\cite{kim-etal-2024-analysis}, Multi-FAct~\cite{shafayat2024multi} + \cite{chataigner2024multilingual}.

% \paragraph{Fact checkers}
% FActScore~\cite{min-etal-2023-factscore}, SAFE~\cite{wei-2024-longform}, VeriScore~\cite{song-etal-2024-veriscore}

% Factcheck-Bench~\cite{wang-etal-2024-factcheck}

\section{Dataset Generation Pipeline}

% \begin{enumerate}
%     \item Class $\rightarrow$ SPARQL query (+geography)
%     \item Entities with (en) Wikipedia pages
%     \item Popularity measures: ...
%     \item Sampling: popularity tier + geography
%     \item Evidence collection: wiki page, wiki search results, linked pages. Add-on: Wikidata triple-based statements. Web search?
%     \item Generate completions by Qwen-2.5-7b-Instruct and llama-3-8b-instruct models.
% \end{enumerate}

Figure~\ref{fig:pipeline} shows the dataset generation pipeline.  
The process begins with defining target classes and extracting class instances from Wikidata using SPARQL queries. 
Depending on the desired class, the query can be very simple (for example, the query to retrieve a list of all rivers contains just one statement \texttt{?x wdt:P31 wd:Q4022}) or more sophisticated. For convenience, the same initial query can collect the locations of the entities and the titles of their corresponding Wikipedia pages. However, when using the public Wikidata query endpoint, complex queries may fail due to timeouts. In this case, additional data can be collected for each entity via the Wikidata API.

After retrieving the basic parameters -- the Wikidata identifiers, entity names (\textit{labels}), and the titles of the corresponding Wikipedia pages in the desired languages -- the popularity scores of the entities are collected. We implemented several proxies for entity popularity described in the literature: 1) a group of parameters based on Wikipedia data, such as the number of pageviews, incoming hyperlinks, edits, and page length; and 2) popularity based on the number of Wikidata triples that the entity is a part of, as either a subject or object. 

Then, we divide the entities into popularity tiers and sample the desired number of entities with predefined popularity and location characteristics. In our preliminary experiments, we found that many rare entities are poorly described in Wikipedia -- the corresponding page contains only one or two sentences, which hinders subsequent evaluation. Therefore, we can introduce the Wikipedia page minimal length and the Wikipedia stub flag as additional inclusion criteria.\footnote{A Wikipedia article is marked as a \textit{stub} by editors if it is considered too short and incomplete.} 

\begin{table}[ht!]
    \centering
    \small
    \begin{tabular}{lrrr} 
                                & \textbf{Rivers}    & \textbf{Disast.}   & \textbf{Cars}   \\ \midrule
    \# Wikidata items           & 426,449   & 6,948       & 12,917  \\ 
    ~~\dots w/ en label         & 330,789   & 5,890       & 11,300  \\ \hline
    \multicolumn{1}{|l}{~~\dots w/ en wiki page}     & 40,914    & 3,706       & \multicolumn{1}{r|}{7,347}   \\ \hline
    ~~~~~~\dots no stubs        & 17,783    & 3,413       & 6,445   \\
    ~~~~~~\dots w/ zh label         & 15,375    & 1,533       & 2,026   \\
    ~~~~~~\dots w/ zh wiki page   & 11,204    & 1,135       & 848    \\
    ~~~~~~~~~~\dots no stubs        & 2,984     & 1,069       & 809    \\ \midrule
    2024 pageviews            & 61.7M & 63.4M & 348.9M\\
%    2024 pageviews            & 61,687,967 & 63,379,378 & 348,864,934\\
    % head                        & 82        & 21         & 268    \\ 
    % torso                       & 810       & 113        & 746    \\
    % tail                        & 39,978    & 3,572       & 6,333   \\ \hline
    % Africa                      & 11,55     & 88         & 15     \\
    % Americas                    & 15,323    & 976        & 1,700   \\
    % Asia/Australia/Oceania      & 5,716     & 844        & 2,401   \\
    % Europe                      & 18,669    & 358        & 5,736   \\
    \end{tabular}
    \caption{Statistics of the \textsc{RiDiC} classes. The framed line corresponds to the \textbf{base class} -- the set upon we collect popularity statistics and define popularity tiers. Entities are sampled from a subset with more stable/complete Wikipedia pages (no stubs, next line).}
    \label{tab:classes_stats}
\end{table}

\begin{table}[ht!]
    \centering
    \small
    \begin{tabular}{lrrr}
            & \textbf{Rivers} & \textbf{Disasters} & \textbf{Cars} \\ \hline
    Head    & 81 (81) & 20 (20) & 100 (77) \\
    Torso   & 200 (150) & 92 (81) & 200 (98) \\
    Tail    & 719 (489) & 888 (622) & 700 (220) \\ \hline
    Africa  & 217 (184) & 18 (8) & 0 (0) \\  
    Americas& 266 (136) & 246 (150) & 233 (67) \\
    AAO     & 264 (171) & 332 (274) & 381 (196) \\ 
    Europe  & 253 (229) & 103 (57) & 371 (129) \\  
    Unknown & 0 (0) & 301 (234) & 15 (3) \\ \hline   
    Total   & 1,000 (720) & 1,000 (723) & 1,000 (395)  \\
    \end{tabular}
    \caption{\textsc{RiDiC} dataset statistics (\# of entities with Chinese Wikipedia pages in parentheses).}
    \label{tab:ridic_statistics}
\end{table}

% For example, when splitting the whole class into three popularity tiers (head-torso-tail) as in the \textsc{RiDiC} dataset, the entities in each part correspond to a third of the total pageviews of the class member in 2024. 
% If geography is additionally used, each quantile is further divided according to the number of pre-defined regions. After that, we can randomly sample entities for the dataset from each part.\footnote{Sometimes, we cannot collect the specified number of entities from each part. For example, we initially planned to have 100 rivers from the ``head'' of the distribution, but only \textbf{XX} most popular rivers by pageviews of the English Wikipedia correspond to a third of the views of all pages about rivers. For more details, see Section \textbf{XXX}.} 
%\footnote{A \textit{stub} is a Wikipedia article considered too short and incomplete. As of 2025, almost half of all Wikipedia's articles are marked as stubs, see \url{https://en.wikipedia.org/wiki/Wikipedia:Stub}} 

Finally, evidence is collected from Wikipedia for the sampled entities. For each entity, we add the corresponding Wikipedia page converted to plain text. Previous studies have shown that using additional knowledge sources makes evaluations more reliable~\cite{song-etal-2024-veriscore,kim-etal-2024-analysis,wei-2024-longform}. 
One can expand the information included in the dataset by using pages that link to the entity's page, as well as the content of pages returned when searching for the entity's name through the Wikipedia search API.
%For tail entities whose pages contain less information, we also collect pages linking to the entity's page. We don't do this for head and torso entities because the number of linked pages can be quite high. Lastly, we add the content of the top-10 pages returned when searching for the entity name through the Wikipedia search API. 
This information makes the dataset self-contained and contributes to the reproducibility of the evaluation results. A richer source of evidence (e.g., web searches) can improve coverage; however, the problem of entity disambiguation would need to be addressed in this case. Using a static knowledge base as a source of evidence may not be the best option if the entities of the selected classes change over time.

\section{\textsc{RiDiC} Dataset}

\begin{table*}[ht]
\footnotesize
    \centering
    \begin{tabular}{l|lll}
         \textbf{Popularity} & \textbf{Rivers} & \textbf{Disasters} & \textbf{Cars} \\ \hline
         Head & Rio Grande & 1958 Lituya Bay earthquake and megatsunami & Xiaomi SU7 \\ 
              & Jordan River & Hurricane Beryl & Honda CR-V\\ 
              & Yangtze & Eruption of Mount Vesuvius in 79 AD & Ferrari Testarossa\\ \hline
         Torso & Grand River (Michigan) & 2024 Spanish floods & Cadillac de Ville series\\ 
                & Medjerda River & 2011 Christchurch earthquake & Mini Countryman \\ 
                & Solo River & Nankai megathrust earthquakes & Mitsubishi Pajero Sport\\ \hline
         Tail & Pequonnock River & 1977 Vrancea earthquake & Cadillac Brougham\\
             & Pastaza River & 2019 East Azerbaijan earthquake & Renault Avantime \\
             & River Yare & 2010 Salang avalanches & Moskvitch 402\\
             %\bottomrule
    \end{tabular}
    \caption{Example entities from the \textsc{RiDiC} dataset (Wikipedia titles).}
    \label{tab:examples}
\end{table*}

We used the proposed pipeline to generate the Rivers/Disasters/Cars (\textsc{RiDiC}) dataset.\footnote{The name refers to \textit{Riddick}, the protagonist of the \textit{Chronicles of Riddick} series, who can see in the dark.
We hope the dataset will improve our understanding of LLMs' capabilities in the dusk area of rare entities.} 
The featured entities belong to three distinct classes: natural objects, natural events, and technical products. We collected the initial set of entities using a SPARQL query corresponding to Wikidata classes \textit{river} (Q4022), \textit{natural disaster} (Q8065), and \textit{automobile model} (Q3231690).\footnote{Note that the \textit{car model} class reflects different manufacturers' naming approaches. For example, each BMW 3 series version is described separately, e.g. \textit{BMW 3 Series (E21)} (Q730915), while all six \textit{Honda CR-V} generations are represented as a single entity (Q255461).}

The classes differ significantly in their structure and size. For instance, there are over 400k river entities on Wikidata, but fewer than 10\% of them are linked to a Wikipedia page.\footnote{We assume this is due to mass imports from an external database or gazetteer.} For disasters and car models, the Wikipedia page requirement reduces the number of objects by a smaller margin, see Table~\ref{tab:classes_stats}. Despite their significant size differences, Rivers and Disasters (40.9k vs. 3.7k) attracted roughly the same amount of attention from Wikipedia users~-- over 60M pageviews in 2024. Cars are much more popular: 7.3k pages received nearly 349M pageviews.

% Imbalance of classes in terms of size and popularity. Note about cars: different granularity, possible errors. Why so many rivers? 
% Next, we identified entities that have a corresponding page in the English Wikipedia. The presence of a Wikipedia page is important in two respects: 1)~only for such entities can we collect Wikpedia-based popularity such as pageviews/hyperlinks and 2)~a Wikipedia page is a ``minimal evidence'' for factuality evaluation of the LLM generations about the entity. Thus, based on two criteria -- class membership and the presence of an English Wikipedia page -- we formed ``core collections'' we later sample dataset entries from. 
We retrieved the locations of entities from Wikidata (for Cars~-- through their manufacturers) and aggregated them into four regions~-- Africa, the Americas, Asia/Australia/Oceania (AAO), and Europe.
% \footnote{When determining the entity's geography, we took the first country/continent from the list returned by Wikidata. Therefore, for objects belonging to transcontinental countries (of which the largest is Russia), the geography may be not precise, but we believe that these cases do not introduce significant noise in our dataset.} 
The geographic distribution of entities is also highly uneven. For example, only 88 (2.4\%) natural disasters and four (0.05\%) cars are attributed to Africa. Wikidata lacks the location of 1,440 (38.9\%) disasters, primarily hurricanes that originate in the ocean and spread across vast territories. 
% See Table~\ref{tab:classes_stats}

% \begin{figure}
%     \centering
%     \includegraphics[width=\linewidth]{images/rivers_popularity_correlations_total.png}
%     \caption{Correlations of different popularity measures in the Rivers class.}
%     \label{fig:popularity_correlation}
% \end{figure}

%\paragraph{Popularity}
% \footnote{\url{https://pageviews.wmcloud.org/massviews/}} 
We used pageviews of an entity's English Wikipedia page in 2024 as the main measure of its popularity.
Based on the collected data, we divided all entities into three popularity tiers: head, torso, and tail. The cumulative number of pageviews for each tier accounts for one-third of the total number of views for the class. The three entity classes exhibit different popularity distributions. E.g., the 81 most popular rivers (0.2\% of the `base class') and the 267 most popular cars (3.6\%) account for one-third of the yearly pageviews in their respective classes.

We also calculated the correlation between a battery of implemented  popularity measures and found that they differ from class to class. In the Rivers, the most skewed collection with a few very popular items, we observed correlations above 0.7 in two groups of scores: 1)~English Wikipedia \#~page edits, pageviews, and \#~inlinks; and 2)~Chinese Wikipedia \#~inlinks and pageviews. The correlation between English and Chinese pageviews is 0.62/0.16/0.60 in Rivers/Disasters/Cars, respectively.
%\footnote{Correlation between English and Chinese pageviews is $0.62$. For example, \textit{Los Angeles River} is \#45 according to EnWiki pageviews and \#6,406 -- to ZhWiki pageviews.} 
In Cars, the strongest correlation is between English Wikipedia page edits and views (0.73), while in Disasters -- between English Wikipedia inlinks and pageviews (0.68). These observations suggest that different popularity proxies are not interchangeable and researchers should carefully choose from the available options.

We collected 1,000 elements from each base class. When possible, we sampled uniformly from four continents and aimed for a 100/200/700 head/torso/tail distribution. Since the heads of the Rivers and Disasters classes contain fewer than 100 entities, we included all of them and sampled additional items from the tail. To ensure that we had enough information to validate LLM responses, we filtered out entities with Wikipedia stubs and pages shorter than 200 characters. This biased the resulting collection slightly toward more popular items; otherwise, we had no data for evaluation. 

In principle, all base class entities can be used for evaluation. However, we believe that 3k entities are close to the optimal size. Note that we obtain 25-30 atomic facts from LLM responses for each entity, which ensures the stability of evaluation. A larger dataset would hinder intensive experiments with different settings and models since evaluating long-form generations is a fairly resource-intensive compared to short answers.\footnote{In our experiments, assessing $\approx$3,000$\times$30 atomic facts from one LLM on a Nvidia RTX 3090 took 36 hours.} 
% Page size threshold, stubs, possible consequences.
% The statistics of the ``core collections'' of the three classes in terms of popularity and geography are presented in the Table~\ref{tab:popularity_geography}. 

% Different popularities, characteristics, correlations. We opted for pageviews, because... Figure~\ref{fig:popularity_correlation}.  

% \paragraph{Geography} 

% \paragraph{Size} 
% Sometimes, we cannot collect the specified number of entities from each part. For example, we initially planned to have 100 rivers from the ``head'' of the distribution, but only \textbf{XX} most popular rivers by pageviews of the English Wikipedia correspond to a third of the views of all pages about rivers. For more details, see Section \textbf{XXX}.
%(see the Appendix \textbf{XXX} with fact checker performance information).

We would like to mention the issue of ambiguity that is often overlooked in other datasets with popularity facet. As we move to less popular items, we observe a higher proportion of entities with the same name. Their Wikipedia titles 
%These entities can be distinguished with additional data In our dataset, we collect both Wikidata labels and Wikipedia page titles, which 
contain disambiguating information in parentheses, e.g. \textit{Colorado River (Argentina)} (there are eight Colorado Rivers on English Wikipedia as of September 2025). If we omit the disambiguation information, there are 3,293/194/169 non-unique names in the base classes corresponding to Rivers/Disasters/Cars. This name ambiguity should be addressed during evaluation, especially when evidence is collected through search.

Finally, we collected information that can serve as evidence for fact verification.
In addition to the Wikipedia page about the entity, we collected the top-10 results returned by the Wikipedia search API using the entity's Wikipedia title as the query and the default parameters.\footnote{\url{https://www.mediawiki.org/wiki/API:Search}} We also collected all Wikipedia pages that link to tail entities because these rare entities typically have limited information on their main Wikipedia page, which makes factual coverage insufficient for reliable verification. Note that, in contrast to search results, linked pages provide  \textit{disambiguated} additional content in the case of namesake entities.

%. \pb{why it is important} 
% incoming links from other wikipedia pages and collected content of all of them which are not just redirect or disambiguation. It was done only for low popularity entities as the coverage of the Wikipedia search results here will be much smaller than in the other ones while incoming links provide chance to gather some more facts about the entity.
When generating the \textsc{RiDiC} dataset, we did not address the problem of the ``ever-greenness'' of facts about the selected entities. However, we believe that rivers, natural disasters and cars are fairly ``stable'' objects, the facts about which do not change much over time.

We added Chinese labels, Wikipedia titles and Wikipedia pages. The dataset can easily be expanded to include other languages. However, the main issue is the quality of factuality \textit{evaluation} in languages other than English~\cite{kim-etal-2024-analysis,shafayat2024multi,chataigner2024multilingual}. 

\textsc{RiDiC} statistics can be found in Table~\ref{tab:ridic_statistics}, examples from different classes and popularity tiers can be seen in Table~\ref{tab:examples}.

\begin{table*}[ht]
    \centering
    \footnotesize
    \begin{tabular}{llrrrrrrrrr}
     & & \multicolumn{3}{c}{\textbf{Rivers}} & \multicolumn{3}{c}{\textbf{Disasters}} & \multicolumn{3}{c}{\textbf{Cars}}  \\ 
     & & \textbf{Llama} & \textbf{Qwen} & \textbf{GPT} & \textbf{Llama} & \textbf{Qwen} & \textbf{GPT} & \textbf{Llama} & \textbf{Qwen} & \textbf{GPT}  \\ \midrule
     \multicolumn{11}{c}{English} \\\midrule
\multirow{2}{*}{Head} 
    & avg. length     & 15.83 & 16.12 & 11.75 & 19.35 & 15.15 & 12.80 & 20.22 & 14.56 & 11.76 \\
    & avg. \# facts   & 32.41 & 33.93 & 29.98 & 28.25 & 30.45 & 25.15 & 34.04 & 29.78 & 26.99 \\ \hline
\multirow{2}{*}{Torso} 
    & avg. length     & 15.27 & 15.75 & 10.64 & 17.91 & 15.65 & 11.82  & 20.40 & 14.80 & 12.27 \\
    & avg. \# facts   & 31.55 & 32.80 & 29.15 & 28.89 & 28.36 & 27.51 & 32.41 & 29.31 & 28.60 \\ \hline
\multirow{2}{*}{Tail} 
    & avg. length     & 14.26 & 14.89 & 9.62 & 16.15 & 15.12 & 10.95 & 19.26 & 14.09 & 11.79 \\
    & avg. \# facts   & 29.85 & 30.91 & 27.18 & 26.72 & 28.52 & 25.09 & 31.44 & 27.69 & 26.70 \\ \hline
\multirow{2}{*}{Total} 
    & avg. length     & 14.59 & 15.16 & 10.00 & 16.37 & 15.17 & 11.07  & 19.58 & 14.28 & 11.88 \\
    & avg. \# facts   & 30.42 & 31.54 & 27.82 & 26.96 & 28.55 & 25.32  & 31.90 & 28.23 & 27.11 \\
    \midrule
    \multicolumn{11}{c}{Chinese} \\\midrule

   \multirow{2}{*}{Head} 
    & avg. length & 13.43 & 13.59 & 10.36 & 11.25 & 11.5 & 9.3 & 15.21 & 11.46 & 11.07 \\ 
    & avg. \# facts & 26.08 & 30.51 & 28.11 & 22.23 & 25.0 & 24.46 & 28.45 & 29.37 & 27.55 \\ \hline
    
    \multirow{2}{*}{Torso} 
    & avg. length & 11.94 & 13.56 & 9.74 & 11.75 & 12.56 & 9.24 & 13.38 & 11.63 & 11.52 \\ 
    & avg. \# facts & 22.60 & 29.67 & 25.35 & 21.91 & 26.81 & 23.45 & 26.27 & 28.95 & 27.43 \\ \hline
    
    \multirow{2}{*}{Tail} 
    & avg. length & 11.76 & 13.20 & 9.08 & 10.82 & 11.91 & 8.87 & 13.59 & 11.17 & 11.29 \\ 
    & avg. \# facts & 21.69 & 29.62 & 22.2 & 20.12 & 25.03 & 21.35 & 25.96 & 26.94 & 25.15 \\ \hline
    
    \multirow{2}{*}{Total} 
    & avg. length & 11.93 & 13.31 & 8.45 & 10.91 & 11.96 & 8.73 & 13.71 & 11.29 & 10.31 \\ 
    & avg. \# facts & 22.38 & 29.73 & 23.50 & 20.34 & 25.21 &  21.63 & 26.47 & 27.84 & 26.11 \\ \hline
        
    \end{tabular}
    \caption{Generation statistics: average response length in sentences and average number of extracted atomic facts. Note that in case of Chinese, LLMs are prompted only with entities that have a corresponding Chinese Wikipedia page, see statistics in Table~\ref{tab:ridic_statistics}.}
    \label{tab:generations_stats}
\end{table*}

\begin{table*}[t]
    \centering
    \scriptsize
    \begin{tabular}{p{0.3\textwidth} p{0.42\textwidth} ccc}
    % \begin{tabular}{ccccc}
        \hline
       \textbf{ LLM generation} & \textbf{Atomic facts} & \textbf{1-page} & \textbf{+search} & \textbf{+links} \\
        \hline
            \multirow[t]{5}{0.3\textwidth}{%
Although it never attained hurricane strength, Edouard brought heavy rainfall, localized flooding, and gusty winds across coastal Texas and Louisiana.}
            & Tropical Storm Edouard brought gusty winds. & \cmark & \cmark & \cmark \\
            & Tropical Storm Edouard never attained hurricane strength. & \xmark & \xmark & \cmark \\
            & Tropical Storm Edouard brought localized flooding. & \xmark & \cmark & \cmark \\
            & Tropical Storm Edouard brought heavy rainfall. & \xmark & \cmark & \cmark \\
        \hline
            \multirow[t]{5}{0.3\textwidth}{%
Ultimately, Debby dissipated without causing severe destruction, though it brought heavy rainfall, flash flooding, and localized damage to portions of the Lesser Antilles, Puerto Rico, and the Dominican Republic, where several fatalities were recorded.}
            & Hurricane Debby brought heavy rainfall to Puerto Rico. & \cmark & \cmark & \cmark \\
            & Hurricane Debby dissipated without causing severe destruction. & \xmark & \xmark & \xmark \\
            & Hurricane Debby brought heavy rainfall to the Dominican Republic. & \xmark & \cmark & 
            \cmark \\
            & Hurricane Debby brought flash flooding to portions of the Lesser Antilles & \xmark & \cmark & \cmark \\
        \hline
    \end{tabular}
    \caption{GPT-5 responses about two hurricanes, atomic facts extracted, and verification results.}
    \label{tab:generation_example}
\end{table*}

\section{Experiments}
\subsection{Factuality Evaluation}

% The structure of the dataset for evaluation of long-form LLM responses is very simple, but the role of the fact checker used increases significantly. Compare: FActScore~\cite{min-etal-2023-factscore}, LongFact + SAFE~\cite{wei-2024-longform}. The checker parameters, such as what is considered an atomic fact, how to extract it, and the source of evidence used, can significantly affect the factuality assessment for the generated dataset. 

\paragraph{Evaluation Methodology} For factuality evaluation experiments on \textsc{RiDiC}, we adopt a modified version of the FActScore~\cite{min-etal-2023-factscore} fact checking tool. %Furthermore, we evaluated a subset of \textsc{RiDiC} using SAFE~\cite{wei-2024-longform}. 
FActScore implements the LLM-as-a-judge three-step pipeline: (i)~\textit{atomic fact extraction}, (ii)~\textit{evidence retrieval/ranking}, and (iii)~\textit{fact verification} against retrieved evidence. In the fact extraction step, an LLM is prompted to generate self-contained and unambiguous facts that can be verified independently of other facts or the initial text. Then, supporting evidence from an external knowledge source is retrieved and ranked. Finally, each atomic fact is verified and labeled as either `supported' or `not supported'. \textit{Factual precision} is defined as the ratio of supported facts over the total fact count.

\paragraph{Implementation Details} Unlike FActScore, which uses a deprecated InstructGPT~\cite{ouyang2022training} for \textit{fact extraction} and LLama~1~\cite{touvron2023llama} for \textit{fact verification}, we use a single model for both steps. 

Specifically, we apply Llama-3.1-8B~\cite{llama3modelcard} and Qwen2.5-7B~\cite{qwen2.5} to English and Chinese generations, respectively. 
% Specifically, we use Llama-3.1-8B-Instruct\footnote{\url{https://hf.co/meta-llama/Llama-3.1-8B-Instruct}} and Qwen2.5-7B-Instruct\footnote{\url{https://hf.co/Qwen/Qwen-2.5-7B-Instruct}} for English and Chinese generations, respectively. 

Additionally, we re-implement LLM inference using the vLLM library~\cite{kwon2023vllm}, resulting in an app. six times faster inference. While the original FActScore implementation retrieves supporting passages from a local Wikipedia 2023 dump, we use pages collected as a part of our dataset. Each atomic fact is concatenated with the top-5 supporting paragraphs and passed to a verification LLM, which is asked whether the given fact is supported by at least one paragraph.  Specifically, we collected 1) the entity page; 2) Wikipedia search results; and 3) pages pointing to the entity page (for tail entities only). 
Thus, we obtain three factuality scores depending on the evidence type used (single page, +search results, +linked pages).

\begin{table*}[ht]
    \centering
    \footnotesize
    \begin{tabular}{l|rrrrrrrrr}
         & \multicolumn{3}{c}{\textbf{Rivers}} & \multicolumn{3}{c}{\textbf{Disasters}} & \multicolumn{3}{c}{\textbf{Cars}} \\ 
         & \textbf{Llama} & \textbf{Qwen} & \textbf{GPT} & \textbf{Llama} & \textbf{Qwen} & \textbf{GPT} & \textbf{Llama} & \textbf{Qwen} & \textbf{GPT} \\ 
         \midrule
         & \multicolumn{9}{c}{English} \\
         \midrule
    Head               & 0.58 & 0.61 & 0.74 & 0.72 & 0.74 & 0.88 & 0.53 & 0.58 & 0.77 \\
    ~~~~~+search       & 0.55 & 0.57 & 0.66 & 0.70 & 0.67 & 0.76 & 0.52 & 0.54 & 0.67 \\
    Torso              & 0.42 & 0.45 & 0.63 & 0.70 & 0.70 & 0.85 & 0.48 & 0.49 & 0.72 \\
    ~~~~~+search       & 0.43 & 0.45 & 0.60 & 0.65 & 0.61 & 0.72 & 0.48 & 0.46 & 0.64 \\
    Tail               & 0.25 & 0.27 & 0.50 & 0.43 & 0.42 & 0.63 & 0.32 & 0.35 & 0.63 \\
    ~~~~~+search       & 0.29 & 0.31 & 0.53 & 0.48 & 0.46 & 0.62 & 0.35 & 0.36 & 0.59 \\
    ~~~~~+linked pages & 0.29 & 0.29 & 0.52 & 0.49 & 0.47 & 0.59 & 0.35 & 0.35 & 0.55 \\ 
    \midrule
    AAO             & 0.31 & 0.35 & 0.56 & 0.50 & 0.47 & 0.69 & 0.32 & 0.36 & 0.61 \\
    Africa          & 0.28 & 0.31 & 0.55 & 0.56 & 0.56 & 0.78 & --   & --   & --   \\
    Americas        & 0.41 & 0.45 & 0.62 & 0.47 & 0.47 & 0.66 & 0.44 & 0.46 & 0.71 \\ 
    Europe          & 0.24 & 0.25 & 0.45 & 0.55 & 0.50 & 0.74 & 0.41 & 0.42 & 0.69 \\
    \midrule
    Total           & 0.31 & 0.34 & 0.55 & 0.46 & 0.46 & 0.66 & 0.38 & 0.40 & 0.67 \\
    ~~~~~+search    & 0.34   & 0.36 & 0.56   & 0.51 & 0.48 & 0.63 & 0.39 & 0.40 & 0.61 \\
         \midrule
         & \multicolumn{9}{c}{Chinese} \\
         \midrule
    % Head & 0.17 & 0.22 & 0.30 & 0.24 &  0.27 & 0.42 & 0.20  & 0.26 & 0.26  \\ 
    % ~~~~~+search  & 0.22 & 0.25 & 0.37 & 0.31 & 0.27 & 0.51 & 0.23 & 0.28  & 0.28  \\ 
    % % High + 10 linked pages & 0.04 & 0.06 & 0.15   & 0.25& 0.29 & 0.40 & 0.14 & 0.24 & 0.28  \\ 
    
    % Torso & 0.06 & 0.09 & 0.18  & 0.28 & 0.33  & 0.46 & 0.15 & 0.19 & 0.24  \\ 
    % ~~~~~+search & 0.09 & 0.12 &  0.26   & 0.34 & 0.36 & 0.51 & 0.17 & 0.20  & 0.29  \\ 
    % % Medium + 10 linked pages & 0.05 & 0.06 & 0.16  & 0.28 & 0.28 & 0.43 & 0.13 & 0.19 & 0.31 \\ 
    
    % Tail & 0.02  & 0.04 & 0.12   & 0.20 & 0.21  & 0.36 & 0.11 & 0.16  & 0.27  \\ 
    % ~~~~~+search   & 0.04    & 0.05  & 0.16 & 0.26  & 0.26  & 0.40 & 0.13  & 0.18  & 0.28 \\ 
    % ~~~~~+linked pages & 0.04 & 0.08  & 0.20  &  0.30 & 0.31  & 0.44 & 0.12 & 0.17 & 0.28  \\ 
    % \midrule    
    % AAO & 0.07 & 0.10 & 0.19  & 0.23 & 0.23 & 0.35 & 0.13 & 0.19 & 0.26  \\
    % Africa & 0.02 & 0.04  & 0.11  & 0.24 & 0.20 & 0.42  & --- & ---  & ---  \\ 
    % Americas & 0.05 & 0.08  & 0.17   & 0.20 & 0.22 & 0.36  & 0.15 & 0.22  & 0.31 \\ 
    % Europe & 0.04 & 0.05  & 0.14   & 0.25 & 0.23 & 0.39  & 0.14 & 0.19  & 0.26 \\
    % % Asia & 0.07 & 0.11 & 0.19  & 0.23 & 0.23 & 0.35  & 0.13 & 0.19  & 0.26  \\ 
    % % AAO  &  0.03 & 0.04  & 0.10  & 0.20 & 0.21 & 0.30 & --- & ---   & --- \\  
    % \midrule

    % Total  & 0.05 & 0.07  & 0.16  & 0.21 & 0.23  & 0.38 & 0.14 & 0.19  & 0.26 \\ 
    % ~~~~~+search  & 0.07 &  0.09  & 0.21  & 0.27 & 0.27  & 0.42  & 0.16  & 0.20  & 0.28  \\  

    Head & 0.39  & 0.43  & 0.50  & 0.48  & 0.45  & 0.45   & 0.33  & 0.40  &  0.45   \\ 
    ~~~~~+search  & 0.34  & 0.38  & 0.37  & 0.47  & 0.48  & 0.46  & 0.29  & 0.35  & 0.38   \\ 
    % High + 10 linked pages & 0.04 & 0.06 & 0.15   & 0.25& 0.29 & 0.40 & 0.14 & 0.24 & 0.28  \\ 
    
    Torso & 0.26  & 0.31  & 0.39  & 0.41  & 0.42  &  0.47  & 0.26  & 0.29  &  0.41   \\ 
    ~~~~~+search  & 0.17  & 0.20  & 0.24  & 0.29  & 0.31  &  0.34  & 0.20  &  0.21  & 0.28   \\ 
    % Medium + 10 linked pages & 0.05 & 0.06 & 0.16  & 0.28 & 0.28 & 0.43 & 0.13 & 0.19 & 0.31 \\ 
    
    Tail  & 0.15  & 0.18  & 0.35  & 0.27  &  0.27  & 0.39  & 0.18  & 0.23  & 0.40   \\ 
    ~~~~~+search  & 0.09  & 0.10  & 0.15  & 0.12  & 0.10  & 0.15  & 0.15  & 0.16  & 0.27   \\ 
    ~~~~~+linked pages  &  0.13  & 0.16  & 0.27  & 0.12  & 0.11  & 0.15  & 0.15  & 0.18  &  0.26   \\  
    \midrule   

    Americas  & 0.30  & 0.33  & 0.42  & 0.29  & 0.29  & 0.4  & 0.28  & 0.33  & 0.41  \\ 
    Africa  & 0.20  & 0.23  & 0.37  & 0.37  & 0.42  & 0.49 & --  &  -- & --   \\ 
    AAO  & 0.23  & 0.29  & 0.42  & 0.33  & 0.33  & 0.47  & 0.23  & 0.28  & 0.42  \\ 
    Europe  & 0.16  & 0.18  & 0.34  & 0.47  & 0.42  & 0.53  & 0.32  & 0.34  & 0.44   \\ 
    % Asia & 0.07 & 0.11 & 0.19  & 0.23 & 0.23 & 0.35  & 0.13 & 0.19  & 0.26  \\ 
    % AAO  &  0.03 & 0.04  & 0.10  & 0.20 & 0.21 & 0.30 & --- & ---   & --- \\  
    \midrule

    Total & 0.21  & 0.25  & 0.38  & 0.30  & 0.29  & 0.40   & 0.26  & 0.30  & 0.42   \\ 
    ~~~~~+search   & 0.14  & 0.16  & 0.20  & 0.15  &  0.14  &  0.18  & 0.21  & 0.24  & 0.30   \\ 
    
    \bottomrule 
    
    \end{tabular}
    \caption{Factuality evaluation of three LLMs' generations on the RiDiC data.}
    \label{tab:factowl_eval}
\end{table*}

\paragraph{Factuality Evaluation in Chinese} 
Initially, FActScore was validated by comparing its predicted factuality scores with the factuality scores on manually labeled data~\cite{min-etal-2023-factscore}. However, the validation was performed only on English data. To assess FActScore performance in Chinese, we translated the generated facts 
% for 549 generations 
from the FActScore dataset, as well as supporting pages, from English to Chinese using Qwen3-235B. On average, factual precision on the translated data was 7.8\% lower than human judgments (53.9\% vs. 61.7\%). These results highlight that long-form factuality evaluation is still challenging, especially for non-English languages.

\subsection{LLMs' Responses}

% \subsection{Factuality Evaluation Results}

% After the dataset is collected, 
The models Qwen-2.5-7b-Instruct~\cite{qwen2.5}, Llama-3-8b-Instruct~\cite{llama3modelcard}, and GPT-5~\cite{gpt5} are prompted to generate responses for each of the \textsc{RiDiC} entities in two languages~-- English and Chinese. %Generation was done via Transformers framework to ensure highest quality and compatibility with the models. 
% The parameters for sampling are presented in Table~\ref{tab:gen_params} in Appendix. 
% Final dataset statistics are presented in Table~\ref{tab:dataset_statistics}.
%\pb{Descibe Table~\ref{tab:generations_stats}}
% Additionally, we computed summary statistics for model outputs. Following the FActScore methodology \cite{min-etal-2023-factscore}, we extracted atomic facts from each generation and reported the mean number of facts per output. We also segmented the generations into sentences and reported the mean number of sentences per output for each domain.
Despite differences in topic rarity (Head/Torso/Tail), the statistics are broadly similar within each language. Chinese generations are slightly shorter and contain slightly fewer atomic facts than English generations. Llama and Qwen produce comparable counts of sentences and atomic facts, whereas GPT tends to produce fewer of both, see Table~\ref{tab:generations_stats}. Excerpts from GPT-5 responses about hurricanes, extracted atomic facts, and the their evaluations using three different evidence variants can be seen in Table~\ref{tab:generation_example}.

\subsection{Factuality Evaluation Results}

Table~\ref{tab:factowl_eval} shows the factuality scores for LLM generations on the \textsc{RiDiC} dataset, averaged by entity popularity and region. Based on these results, a few observations can be made.

\paragraph{LLMs Hallucinate on Rare Entities More} 
The observed factuality scores are strongly correlated with the entity's popularity. For example, English scores preserve the Head $>$ Torso $>$ Tail ordering for all evaluated LLMs and domains. In particular, Llama shows about 2x factuality drops from 0.58/0.72/0.53 on \textit{Head} entities to 0.25/0.43/0.32 on \textit{Tail} for Rivers/Disasters/Cars, respectively. The same holds for Qwen generations in English. GPT-5 is slightly more robust with respect to entity popularity, with 0.74-0.88 and 0.50-0.63 score ranges on \textit{Head} and \textit{Tail}, respectively. 

Similarly to the English evaluation, GPT-5 is consistently more factually accurate than the smaller Qwen and Llama models. The model also shows a much lower factuality gap between frequent and rare entities in Chinese. In contrast, Llama and Qwen have about 2x gap between frequent and rare entities. For instance, Llama's performance falls from 0.39/0.48/0.33 on \textit{Head} entities to 0.15/0.18/0.35 on \textit{Tail} for Rivers/Disasters/Cars. Overall, the results suggest that smaller LLMs struggle to memorize long-tail facts more severely due to their limited parametric capacity.
% seem to struggle with factuality regardless of entity popularity.
% Surprisingly, all three models show higher scores on \textit{Torso} than on \textit{Head} entities for disasters. However, Llama  and Qwen  have an $\sim$8x (0.17 vs. 0.02) and a $\sim$5x (0.22 vs 0.04) decrease in factuality, respectively, on Rivers. 

% In Chinese, Llama and Qwen seem to struggle with factuality regardless of entity popularity.
% Surprisingly, all three models show higher scores on \textit{Torso} than on \textit{Head} entities for disasters. However, Llama  and Qwen  have an $\sim$8x (0.17 vs. 0.02) and a $\sim$5x (0.22 vs 0.04) decrease in factuality, respectively, on Rivers. Overall, the results suggest that LLMs struggle to memorize long-tail facts due to their limited parametric capacity and the imbalanced entity distribution in their pre-training data.

\paragraph{Low Factuality in Chinese}
% Factuality scores for English generations are considerably higher than those for Chinese across all models, domains, and popularity tiers.
%Although we performed the evaluation on the same domains and sets of entities, 
% In Chinese, all three models show lower factuality scores compared to English evaluation. This observation can be explained by either the language gap or the incompleteness of information within the Chinese Wikipedia.

In Chinese, all three models show lower factuality scores compared to English evaluation. Particularly, the total factuality scores for English range from 0.31 to 0.67 across domains, while for Chinese generations, they lie within the range of 0.21 to 0.42. Additional context from  Wikipedia search seems to further mislead the evaluation by introducing lowly relevant textual passages causing a total factuality drop ranging from  0.07 to 0.18 (Llama and GPT evaluation on Rivers).  

We hypothesize that this gap is caused by the following challenges. First, current LLMs tend to perform worse in non-English languages due to less intensive pre-training and fewer high-quality linguistic resources, leading to weaker generation capabilities and lower factual precision in Chinese~\cite{DBLP:journals/corr/abs-2404-11553}. 

Second, the gap can be attributed to the limited reference data available on Chinese Wikipedia, which reduces the reliability and scope of evidence for fact verification~\cite{DBLP:conf/acl/HeLLTWHBGHZLSZS25}. Overall, the findings indicate current challenges in generating and verifying accurate long-form output in non-English languages and reveal the need for improved multilingual LLM capabilities and factuality evaluation methods.

\paragraph{Factuality is Sensitive to Domain} Factuality scores vary notably across the three domains.
\textit{Disasters} and \textit{Cars} receive the highest factuality scores from all LLMs, and \textit{Rivers} receive the lowest factuality scores. Therefore, an LLM's ability to generate factually accurate texts depends on the target domain, but the domain complexity is generally consistent across different LLMs.
% suggesting that LLMs generate more accurate and reliable information on this domain. 

\paragraph{LLMs Exhibit no Clear Geographic Bias} The effect of an entity's location on the factual accuracy is mixed. English responses about American Rivers and Cars are more accurate, but responses about Disasters show a different picture (note that \textsc{RiDiC} contains no cars and very few disasters attributed to Africa, so due to the sample size, the final scores may be skewed). It is possible that location's impact is more significant at the level of individual countries, as shown by~\cite{shafayat2024multi}, rather than across entire continents.

\paragraph{Smaller LLMs are Less Accurate}

GPT-5 achieves the highest factuality scores across all domains and popularity levels, indicating superior factual precision in long-form generation. The Llama and Qwen models exhibit a similar level of factual hallucinations, which suggests that models of comparable size have similar capacities for memorizing factual knowledge. 
% behaviour in terms of factuali

% Among the evaluated large language models, GPT-5 consistently achieves the highest factuality scores across all domains and popularity tiers, indicating superior factual precision in long-form generation. Qwen2.5-7B-Instruct models typically perform at a medium level, while Llama models yield the lowest factuality scores. This ranking holds true for both English and Chinese generations, demonstrating GPT’s stronger overall factuality capabilities compared to Qwen and Llama.

% Factuality performance varies notably across the three domains: rivers, cars, and natural disasters. Natural disasters receive the highest factuality scores from all LLMs, suggesting that LLMs generate more accurate and reliable information on this domain. Cars have moderate factuality scores, and rivers receive the lowest factuality scores among the three domains. This indicates domain-specific differences in the factual knowledge or generation quality of LLMs.

\paragraph{Richer Evidence's Harm and Benefit}
% Helps Non-English Verification
% On average, using Wikipedia search results and linked pages for rare entities does not increase factual precision in English. Additional context decreases the scores for \textit{Head} and \textit{Torso} as the target entity's  page already seems to provide sufficient context for factuality assessment. Conversely, Chinese evaluations decreases when adding linked and search pages. Possibly, the observation can be caused by poorer evidence ranking in Chinese. 
% Using Wikipedia search results and linked pages for \textit{Tail} entities increases factual precision for rare entities in English  for all three models on \textit{Rivers} and for \textit{Qwen} and \textit{Llama} models on \textit{Disasters}. However, additional context decreases the scores for \textit{Head} and \textit{Torso} as the target entity's  page already seems to provide sufficient context for factuality assessment. Despite less comprehensive page content, Chinese evaluations decrease when adding linked and search pages for all three frequency groups. Possibly, the observation can be caused by poorer evidence ranking in Chinese which results in misleading contexts being retrieved at high rank. 

For English, incorporating Wikipedia search results and linked pages enhances factual precision for \textit{Qwen} and \textit{Llama} models on \textit{Tail} entities. However, this additional context reduces performance for \textit{Head} and \textit{Torso} entities, as their primary pages already appear to provide sufficient context for factuality assessment. In contrast, Chinese performance drops across all frequency groups when supplementary pages are added, despite the generally less comprehensive nature of Chinese Wikipedia content. This observation may be caused by poorer evidence ranking in Chinese, resulting in misleading contexts being retrieved at high ranks.

\begin{table}[h]
    \centering
    \small
    \begin{tabular}{llrrr}
      \textbf{Model} & \textbf{Tier} & \textbf{\#} & \textbf{RMSE} (km) & \textbf{MAPE} \\ \hline
      Llama & Head  & 81    & 84 & 6.62 \\
            & Torso & 187   & 216 & 10.51 \\ 
            & Tail  & 496   & 223 & 63.38 \\ \hline
      Qwen  & Head  & 78    & 86 & 2.53 \\
            & Torso & 187   & 143 & 21.46 \\
            & Tail  & 507   & 256 & 80.17 \\ \hline
      GPT5  & Head  & 76    & 51 & 10.11 \\
            & Torso & 179   & 67 & 3.76 \\
            & Tail  & 420   & 99 & 16.90 \\
            % \hline
    \end{tabular}
    \caption{Errors in river lengths: English generations vs. Wikidata. 
    Note that the \textit{relative} error (MAPE) is smaller in case of popular rivers -- they are in general longer.
    }
    \label{tab:rmse_rivers}
\end{table}

\begin{figure}[t!]
    \centering
    \includegraphics[width=0.99\linewidth]{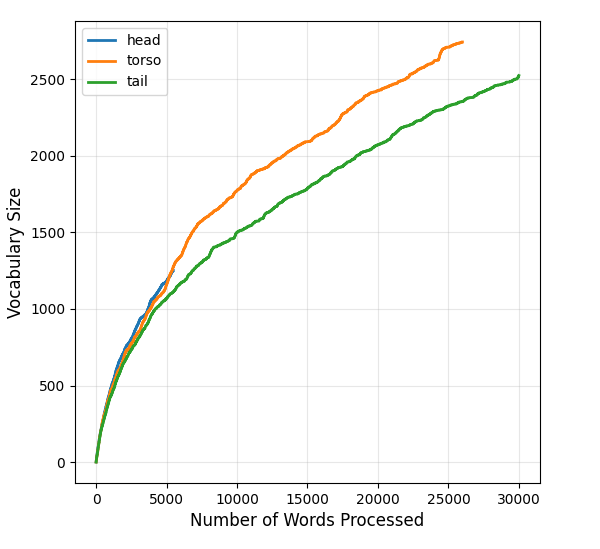}
    \caption{Vocabulary growth in Qwen's generations about disasters in different popularity tiers.}
    \label{fig:heaps_qwen_disasters}
\end{figure}

\paragraph{Targeted Evaluation} In addition to the FActScore-based evaluation, we  conducted an evaluation focused on a single attribute~-- river length. 
%(we also evaluated the dates of natural disasters and vehicle lengths, but we do not report the results here due to space constrains). 
To this end, we extracted river lengths from Wikidata (property \textit{P2043}), detected the lengths in the LLM responses using a regular expression, and compared them. In some cases, we observed discrepancies between Wikidata and Wikipedia. For example, as of September 2025, Wikipedia indicated the length of the Chicago River as 156 miles, whereas it was erroneously indicated as 1.6 miles in Wikidata. This observation again highlights the dependence of factuality evaluation on the knowledge source used.

The results are presented in Table~\ref{tab:rmse_rivers}. This approach demonstrates that at least some aspects of long-form generations allow for more fine-grained evaluations than binary decisions on atomic facts. 

% \begin{figure*}[t]
%   \centering
%   % one image spanning both columns
%   % \includegraphics[width=\textwidth]{images/your_wide_plot.pdf}

%   % or three side-by-side PDFs across the full page width:
%   \begin{subfigure}[t]{0.32\textwidth}
%     \centering
%     \includegraphics[width=\linewidth]{images/gpt_barplot.pdf}
%     \caption{GPT}
%   \end{subfigure}\hfill
%   \begin{subfigure}[t]{0.32\textwidth}
%     \centering
%     \includegraphics[width=\linewidth]{images/qwen_barplot.pdf}
%     \caption{Qwen}
%   \end{subfigure}\hfill
%   \begin{subfigure}[t]{0.32\textwidth}
%     \centering
%     \includegraphics[width=\linewidth]{images/llama_barplot.pdf}
%     \caption{Llama}
%   \end{subfigure}

%   \caption{Supported atoms by relative position.}
%   \label{fig:wide-bars}
% \end{figure*}

\paragraph{Vocabulary Richness} We investigated another aspect of LLM generations: their lexical diversity. To this end, we measured vocabulary growth (the number of unique words) as a function of text length (concatenated LLM responses) across classes and popularity tiers -- the relationship described by the Heaps' law~\cite{heaps1978information}. Our results suggest that generations about less popular entities have lower lexical diversity. This effect is more pronounced in the case of Disasters (see Figure~\ref{fig:heaps_qwen_disasters}) and Rivers, but is almost nonexistent in Cars. Among the three LLMs, GPT exhibits the highest lexical diversity, followed by Qwen and Llama.

\section{Conclusion}
This paper introduces a highly configurable, multilingual pipeline designed to generate datasets with controlled entity popularity distributions. This pipeline is intended to facilitate the long-form factuality evaluation of LLMs. Our approach allows for the systematic sampling of entities from various domains, geographic regions, and popularity tiers. We also introduce \textsc{RiDiC}, a dataset comprising 3,000 entities from three domains.

Our experimental results demonstrate that current LLMs exhibit significant variability in factual precision based on entity popularity, language, and domain, demonstrating notable weaknesses with long-tail entities and in non-English scenarios. Thus, the \textsc{RiDiC} dataset and tools provide valuable resources for advancing automatic factuality assessment and accelerating progress in trustworthy language generation. 
The study also revealed challenges in evaluating non-English generations.
Future research would benefit from extending these methods to a more diverse set of domains, languages, and LLMs.

\section*{Limitations}

While the \textsc{RiDiC} dataset and generation pipeline offer a robust framework for evaluating long-form factuality in large language models, several limitations must be acknowledged. First, experiments are conducted on only three LLMs (two of which are of a similar size) and in only two languages: English and Chinese. 

Second, factuality evaluations for non-English languages are less reliable due to the scarcity of high-quality reference data, as well as the lower performance of current LLMs in these languages. 

Third, excluding Wikipedia stubs and short articles results in a dataset that is modestly biased towards popular entities, thereby underrepresenting the long tail of entity popularity.

Furthermore, as English Wikipedia pageviews are used as a popularity signal, the popularity distribution may be biased. Finally, reliance on Wikipedia as a primary source may limit the range of evidence covered. Future work should address these issues by improving multilingual evidence collection, refining approaches to resolving ambiguity, and extending the pipeline to incorporate more diverse and up-to-date knowledge sources.

% Since December 2023, a "Limitations" section has been required for all papers submitted to ACL Rolling Review (ARR). This section should be placed at the end of the paper, before the references. The "Limitations" section (along with, optionally, a section for ethical considerations) may be up to one page and will not count toward the final page limit. Note that these files may be used by venues that do not rely on ARR so it is recommended to verify the requirement of a "Limitations" section and other criteria with the venue in question.

% \section*{Acknowledgments}

% Bibliography entries for the entire Anthology, followed by custom entries
%\bibliography{anthology,custom}
% Custom bibliography entries only
% \bibliography{anthology,latex/custom}

\section{Bibliographical References}\label{sec:reference}

\bibliographystyle{lrec2026-natbib}
\bibliography{lrec}

% \section{Language Resource References}
% \label{lr:ref}
% \bibliographystylelanguageresource{lrec2026-natbib}
% \bibliographylanguageresource{languageresource}

% \appendix

% \section{FActScore on Chinese Translations}
% Evaluation of FActScore's atomic facts translated to Chinese, is presented in Table~\ref{tab:factowl_on_chinese_translations}.
% \include{latex/tables/appx_chinese_translations}

% \section{Appendix}
% \label{sec:appendix}

% \begin{table}[htbp]
%     \centering
%     \small
%     \begin{tabular}{lr}
%         \toprule
%         \textbf{Parameter} & \textbf{Value} \\
%         \midrule
%         max\_new\_tokens & 512 \\
%         do\_sample & True \\
%         temperature & 0.7 \\
%         early\_stopping & False \\
%         num\_beams & 1 \\
%         top\_k & 50 \\
%         top\_p & 1.0 \\
%         repetition\_penalty & 1.0 \\
%         diversity\_penalty & 0.0 \\
%         \bottomrule
%     \end{tabular}
%     \caption{Generation Hyperparameters for Text Generation}
%     \label{tab:gen_params}
% \end{table}
\end{document}